\PassOptionsToPackage{table}{xcolor}
\documentclass{article}
\usepackage{iclr2027_conference,times}

\usepackage{amsmath,amsfonts,bm}

\def\eqref#1{equation~\ref{#1}}

\def\1{\bm{1}}

\DeclareMathAlphabet{\mathsfit}{\encodingdefault}{\sfdefault}{m}{sl}
\SetMathAlphabet{\mathsfit}{bold}{\encodingdefault}{\sfdefault}{bx}{n}

\usepackage{url}
\usepackage{xcolor}
\usepackage{booktabs}
\usepackage{multirow}
\usepackage{array}
\usepackage{tabularx}
\usepackage{graphicx}
\usepackage{float}
\usepackage{placeins}
\usepackage{wrapfig}

\usepackage{hyperref}  

\definecolor{igsdCaseGray}{HTML}{F2F2F2}
\definecolor{igsdCaseMuted}{HTML}{656565}
\definecolor{igsdCaseStudent}{HTML}{246695}
\definecolor{igsdCaseTeacher}{HTML}{975B21}
\definecolor{igsdCaseGood}{HTML}{226F4A}
\definecolor{igsdCaseBad}{HTML}{A83140}
\newcommand{\igsdStudent}[1]{\textcolor{igsdCaseStudent}{\textbf{#1}}}
\newcommand{\igsdTeacher}[1]{\textcolor{igsdCaseTeacher}{\textbf{#1}}}
\newcommand{\igsdCorrect}[1]{\textcolor{igsdCaseGood}{\textbf{#1}}}
\newcommand{\igsdIncorrect}[1]{\textcolor{igsdCaseBad}{\textbf{#1}}}

\title{IGSD: Environment-Verified Hindsight Self-Distillation for Search Agents}

\author{%
\textbf{Angqing Jiang}\textsuperscript{1} \quad
\textbf{Gaoming Zhang}\textsuperscript{1} \quad
\textbf{Chaoqun Zhang}\textsuperscript{2} \quad
\textbf{Jianchun Song}\textsuperscript{3} \\
\textbf{Liyuan Kong}\textsuperscript{3} \quad
\textbf{Kena Qi}\textsuperscript{3} \quad
\textbf{Wei Lin}\textsuperscript{3} \quad
\textbf{Defu Lian}\textsuperscript{1}\thanks{Corresponding author.} \\[4pt]
{\mdseries\small
\textsuperscript{1}University of Science and Technology of China \quad
\textsuperscript{2}Tsinghua University \quad
\textsuperscript{3}Meituan} \\[3pt]
{\mdseries\small\texttt{philipgaq@mail.ustc.edu.cn \quad
zzzgm@mail.ustc.edu.cn}} \\
{\mdseries\small\texttt{liandefu@ustc.edu.cn \quad
zhangcq26@mails.tsinghua.edu.cn}} \\
{\mdseries\small\texttt{\{songjianchun,kongliyuan,qikena,linwei31\}@meituan.com}}
}

\begin{document}

\maketitle

\begin{abstract}
On-policy self-distillation densifies agent training without external teachers:
a policy conditioned on privileged hindsight provides step-level guidance for
its own unprivileged rollouts. For search agents, however, hindsight can make
the teacher prefer a query that does not improve retrieval from the student's
state. Existing methods either distill this preference directly or filter it
with model-internal scores, but neither strategy verifies the query's executed
retrieval consequence. We propose \textbf{Information-Gain-Gated
Self-Distillation (IGSD)}, which verifies on-policy token proposals with
environment feedback before distilling them. Treating each query token as a
micro-action, IGSD completes the teacher's token proposal and the student's
sampled token into matched queries and executes both from the same failed
state with the same retriever. Shared counterfactual controls account for
query-conditioned shifts in answer
likelihood, so their difference, the \emph{executed paired information gain},
provides a relative utility contrast for the retrieved documents. IGSD uses
this contrast as a positive-only soft weight for candidate-pair distillation,
while leaving the
GRPO objective unchanged and confining verification to training. Across seven
single-hop and multi-hop QA benchmarks, IGSD reaches macro-average exact-match
accuracies of 42.8\% and 47.0\% with 3B and 7B policies, respectively, without
inference-time verification. These results support
environment-verified hindsight as an effective approach to reliable
action-level supervision for search agents.
\end{abstract}

\section{Introduction}
\label{sec:intro}

Search-augmented reasoning agents solve information-seeking tasks by
interleaving internal reasoning with external retrieval operations~\citep{yao2023react,jin2025searchr1,chen2025research}.
Training frameworks such as Search-R1~\citep{jin2025searchr1} optimize these
trajectories with reinforcement learning under trajectory-level outcome
rewards~\citep{shao2024deepseekmath,guo2025deepseekr1}. Because the reward
depends only on the final answer, an informative query and a redundant or
misleading query can receive the same trajectory-level credit, leaving
intermediate search decisions without fine-grained supervision.

To densify intermediate supervision, recent work introduces \emph{hindsight
self-distillation} (HSD)~\citep{ma2026sdsearch,lu2026sdar}. The policy first
rolls out without privileged context, while a self-teacher conditioned on
hindsight information evaluates or revises intermediate steps and distills
supervision back into the unprivileged policy. For search agents, however, a
query that proved useful in a successful sibling may be redundant or
misleading at the student's failed state: the trajectory may already contain
the relevant evidence, or may lack the bridge entity required by the query.
Hindsight preference is therefore not sufficient evidence that the proposed
query will improve retrieval from the student's current state.

Existing methods can filter hindsight supervision with model-internal
signals, such as the detached teacher--student likelihood gap used by
SDAR~\citep{lu2026sdar}, yet any such score reflects the model's
preference, not whether the proposal actually brings the new evidence the
current student needs. This creates a \emph{teacher-supervision reliability
gap}: how to verify, before distillation, whether teacher guidance
genuinely helps. In search agents, however, every search step naturally
yields environment feedback, which makes this question answerable by
execution rather than by introspection.

We propose \textbf{Information-Gain-Gated Self-Distillation (IGSD)}, which
grounds hindsight guidance in environment-derived utility rather than model
preference. Working on-policy at the states the student actually visited,
the hindsight teacher proposes token-level alternatives; whenever a
proposal diverges from the student's sampled action, IGSD lets the student
policy complete both candidates into matched queries and executes them
with the same retriever. Each execution is scored by how much its
retrieved documents raise the policy's answer likelihood, net of a shared
counterfactual control, and their difference, the \emph{executed paired
information gain}, is an environment-derived utility contrast at the
granularity of a single action. IGSD uses it as a positive-only soft
weight on candidate-pair distillation, so helpful proposals receive
stronger supervision while non-improving ones are omitted. The GRPO
objective remains unchanged, and verification is confined to training.

Experiments across seven single-hop and multi-hop QA benchmarks show that
IGSD achieves the highest macro-average exact-match accuracy among the
methods, reaching 42.8\% with the 3B policy and 47.0\% with the
7B policy, with no inference-time verification.
Our code is available at
\url{https://github.com/PhilipGAQ/igsd_code}.

Our main contributions are as follows:
\begin{itemize}
    \item \textbf{Environment-verified hindsight distillation.} We introduce IGSD, which converts hindsight teacher suggestions into local candidate actions and verifies their retrieval consequences from the student's state before distillation.
    \item \textbf{Paired utility gating.} We define executed paired information gain from matched teacher and student query executions with shared counterfactual controls, and use it as a positive-only soft weight on the candidate-pair distillation loss while leaving the GRPO objective unchanged.
    \item \textbf{Empirical and mechanistic analysis.} We validate the effectiveness of IGSD across seven single-hop and multi-hop QA benchmarks at both 3B and 7B scales, and an extensive mechanistic analysis shows stronger observed alignment with terminal outcomes than the tested model-internal signals.
\end{itemize}

\section{Related Work}
\label{sec:related}

\paragraph{Search-agent RL.}
Search-R1~\citep{jin2025searchr1} trains language models to interleave
reasoning and retrieval with a trajectory-level outcome reward. ReSearch~\citep{chen2025research}
extends this setting to multi-turn reasoning and search control. Other work
introduces denser step- or action-level signals, including IGPO~\citep{wang2025igpo}
and GiGPO~\citep{feng2025gigpo}. IGSD differs by adding a paired,
environment-executed comparison of candidate search actions from the same
local state, while preserving the trajectory-level objective and introducing
the action-level signal only through a soft distillation weight.

\paragraph{Hindsight self-distillation.}
SD-Search~\citep{ma2026sdsearch} applies on-policy hindsight
self-distillation to search agents, but does not perform per-candidate
retrieval verification. Search-E1~\citep{liang2026searche1} alternates GRPO
updates with privileged-context self-distillation. SDAR~\citep{lu2026sdar}
gates per-token distillation with a detached teacher--student likelihood gap,
which measures an internal preference rather than the utility of retrieved
evidence. Skill-SD~\citep{wang2026skillsdskillconditionedselfdistillationmultiturn} conditions privileged
self-distillation on skills for multi-turn agents. IGSD complements these
approaches by executing teacher-induced and student-induced queries from the
same state and using their paired environment-derived utility to decide
whether distillation is applied.

\paragraph{Information gain as supervision.}
IG-Search~\citep{liang2026igsearch} and IGPO~\citep{wang2025igpo}
introduce information gain as an RL signal. SIGHT~\citep{zhong2026sight}
uses it for observation-level de-duplication, reflection, and branching,
while SmartSearch~\citep{wen2026smartsearch} applies it to query verification
and refinement. CRAFT~\citep{meng2026craft} and LOTAPO~\citep{zhu2026lotapo}
estimate token- or process-level credit from sibling rollouts. IGSD instead
uses paired executed gain only to weight auxiliary distillation, without
changing the policy-gradient objective.

\section{Method}
\label{sec:method}

\begin{figure}[t]
\centering
\includegraphics[width=0.9\textwidth]{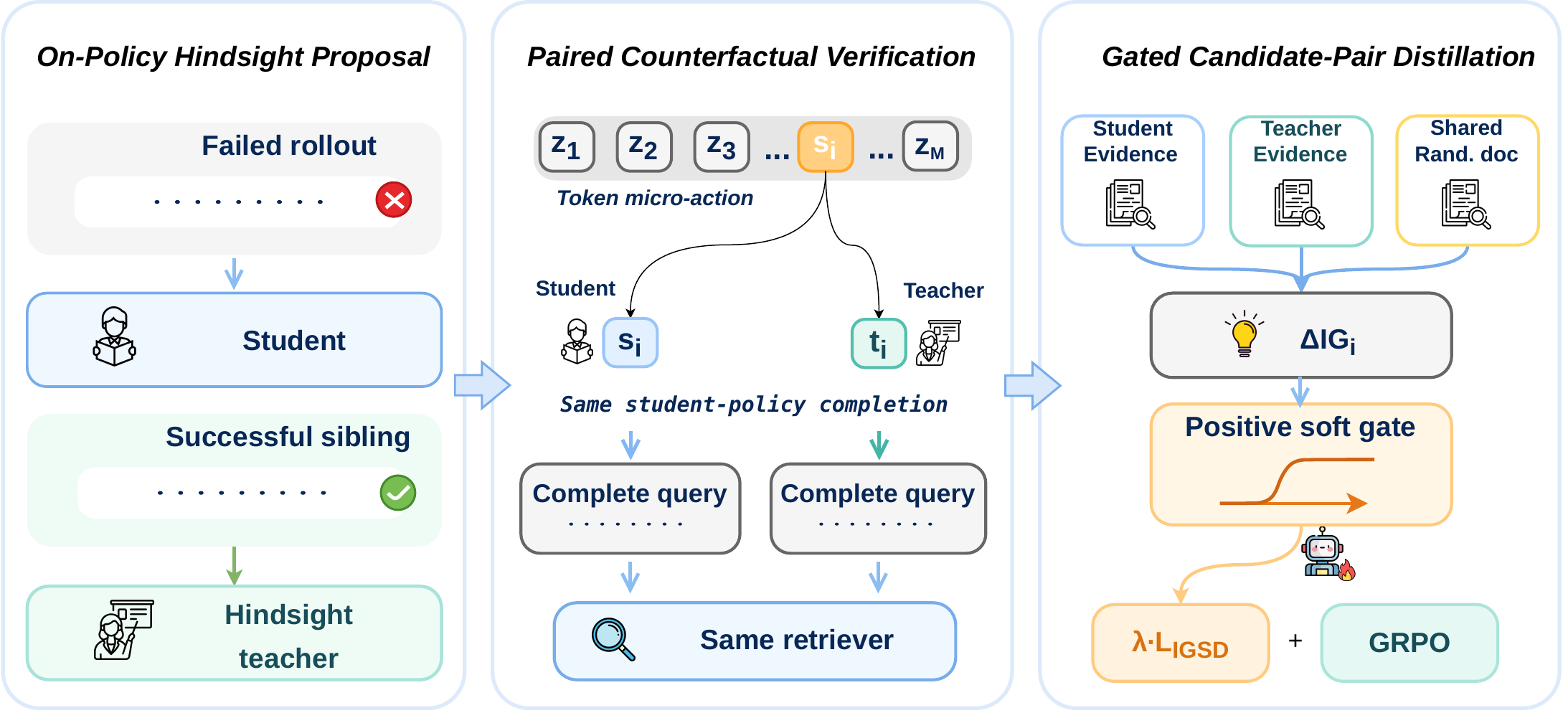}
\caption{Overview of IGSD. Left: privileged hindsight produces a teacher
proposal for a failed student rollout. Middle: the teacher and sampled student
tokens are independently completed by the same frozen student policy into
matched queries and evaluated with the same retriever. Right: the resulting environment-derived
utility contrast $\Delta\mathrm{IG}_i$ gates candidate-pair distillation
together with the unchanged GRPO objective.}
\label{fig:igsd-overview}
\end{figure}

IGSD augments GRPO with environment-verified action-level supervision. As
shown in Figure~\ref{fig:igsd-overview}, a hindsight-conditioned teacher
proposes local alternatives to a failed student trajectory, IGSD executes
the resulting candidate queries under matched conditions, and the utility
contrast measured on the retrieved evidence gates candidate-pair
distillation while leaving the GRPO objective unchanged.
Section~\ref{sec:problem-setup} formalizes the search-agent setting;
Section~\ref{sec:setting} constructs hindsight-guided candidates;
Section~\ref{sec:verify} defines the paired environment verification that
produces the executed information gain; and Section~\ref{sec:distill}
presents the gain-gated distillation objective.

\subsection{Problem Setup}
\label{sec:problem-setup}

\paragraph{Search-agent trajectories.}
Given a question $x$, a search policy $\pi_\theta$ generates a multi-turn
trajectory $\tau$ that interleaves reasoning, search actions, and retrieved
observations before producing a terminal answer. After standard answer
normalization, the answer evaluator returns a scalar terminal outcome score
$r(\tau)$ rather than a strict binary exact-match label; partially matching
answer strings can therefore receive nonzero score. We use its
answer-correctness signal to label trajectories as successful or failed, while
retaining the scalar score for sibling ranking and the hindsight hint. GRPO optimizes this terminal score
over rollouts for the same question, but does not identify which intermediate
search decisions improve retrieved evidence. IGSD derives such local
supervision while leaving the GRPO reward and advantage estimator unchanged.

\subsection{Hindsight-Guided Candidate Construction}
\label{sec:setting}

\paragraph{Student trajectory.}
IGSD constructs auxiliary supervision for a failed trajectory whose rollout
group contains a successful sibling and at least one executable search action.
It selects the last valid search action for supervision. Let $h_S$ denote the
prefix before this action and $z=(z_1,\ldots,z_M)$ its serialized tool call.
The tokenizer positions inside the search-query field are
$\mathcal{I}_Q=\{k_1,\ldots,k_N\}$, excluding tool-call syntax and other
fields. The sampled student token at position $k_i$ is $s_i=z_{k_i}$.

\paragraph{Hindsight teacher.}
A selected successful sibling provides a search-strategy hint containing its
search queries $Q^+$ and outcome score $r^+$, denoted by
$\mathcal{H}^+=(Q^+,r^+)$. The hint excludes the sibling's terminal answer,
reasoning, retrieved documents, and tool observations. The teacher uses the
same frozen policy checkpoint as the rollout policy but is conditioned on the
student trajectory prefix and the hindsight hint. When multiple successful
siblings are available, we select the one with the highest terminal reward,
breaking ties by rollout order:
\begin{equation}
h_T=\mathcal{P}_T(x,h_S,\mathcal{H}^+),
\label{eq:hindsight_context}
\end{equation}
Evaluating teacher and student along the same sampled action prefix gives
\begin{equation}
p_i^T(v)=\pi_{\theta_{\mathrm{old}}}(v\mid h_T,z_{<k_i}),
\qquad
p_i^S(v)=\pi_{\theta_{\mathrm{old}}}(v\mid h_S,z_{<k_i}).
\label{eq:teacher_student_distributions}
\end{equation}
The two distributions share the decision state and differ only in the
presence of the hindsight hint.

\paragraph{Candidate micro-actions.}
We treat each query token as a micro-action. At position $k_i$, the teacher
proposes $t_i=\arg\max_v p_i^T(v)$: where $t_i=s_i$ the teacher endorses
the student's action, and each disagreement defines a candidate pair
$\mathcal{S}_i=\{t_i,s_i\}$ to be verified on-policy, by executing both
actions at the exact state visited by the student's own rollout and
comparing the evidence they retrieve (Section~\ref{sec:verify}).

\subsection{Paired Environment Verification}
\label{sec:verify}

\paragraph{Matched query completion.}
IGSD converts each candidate pair into two executable queries by fixing the
candidate at position $k_i$ and greedily completing the remainder with the
frozen student policy, without the hindsight hint:
\begin{equation}
Q_i^T=\operatorname{Complete}_{\pi_{\theta_{\mathrm{old}}}}
  (h_S,z_{<k_i}\oplus t_i),
\qquad
Q_i^S=\operatorname{Complete}_{\pi_{\theta_{\mathrm{old}}}}
  (h_S,z_{<k_i}\oplus s_i).
\label{eq:matched_branches}
\end{equation}
The two branches share the state, prefix, continuation policy, decoding
procedure, and retriever; only the fixed token at $k_i$ differs. The resulting
comparison measures the downstream utility of a local token intervention
under the shared continuation policy. We call them the teacher-induced and
student-induced queries, with retrieved document sets $D_i^T$ and $D_i^S$.

\paragraph{Document utility and counterfactual control.}
IGSD scores document set $D$ by the frozen policy's mean
reference-answer token log-likelihood. For branch $b\in\{T,S\}$ and answer
aliases $\mathcal{A}_i$, define
\begin{equation}
L_i^b(D;\mathcal{A}_i)
=\max_{a\in\mathcal{A}_i}
\log\pi_{\theta_{\mathrm{old}}}(a\mid h_S,Q_i^b,D).
\label{eq:answer_evidence}
\end{equation}
Here, the log-likelihood of an alias denotes the mean over its encoded answer
tokens. The maximum lets any valid surface form of the reference answer carry
the score. The score $L_i^b$, however, also responds to the query itself: a
query's phrasing can shift the policy's answer likelihood even when its
retrieved evidence is unchanged. To reduce this query-dependent component as
much as possible while keeping the comparison focused on retrieved evidence,
IGSD scores every branch against a shared counterfactual control: for each
candidate pair it draws $C$ counterfactual document sets
$\widetilde D_{ij}$ at random from the
corpus, each matched in size and format to the retrieved documents.

\paragraph{Executed paired information gain.}
The executed paired information gain compares the controlled document
utilities of the two branches:
\begin{equation}
\Delta\mathrm{IG}_i
=\frac{1}{C}\sum_{j=1}^{C}
\Big[
\big(L_i^T(D_i^T;\mathcal{A}_i)
     -L_i^T(\widetilde D_{ij};\mathcal{A}_i)\big)
-\big(L_i^S(D_i^S;\mathcal{A}_i)
     -L_i^S(\widetilde D_{ij};\mathcal{A}_i)\big)
\Big].
\label{eq:paired_ig}
\end{equation}
A positive $\Delta\mathrm{IG}_i$ indicates that the evidence retrieved by the
teacher-induced query contributes more answer support than the evidence
retrieved by the student's own query. This environment-derived utility
contrast weights the auxiliary distillation loss but does not enter the GRPO
reward or advantage estimator.

\subsection{Gain-Gated Candidate-Pair Distillation}
\label{sec:distill}

\paragraph{Positive-only utility gate.}
The executed gain determines whether a candidate pair contributes to
distillation and how strongly it is weighted. IGSD defines the detached soft
gate
\begin{equation}
w_i=\operatorname{sg}\!\left[
\max\!\left(
0,\,
\tanh\!\left(\frac{\beta}{2}\Delta\mathrm{IG}_i\right)
\right)
\right],
\label{eq:utility_gate}
\end{equation}
where $\operatorname{sg}$ denotes stop-gradient and $\beta$ controls the
sharpness; candidate pairs with non-positive gain receive zero auxiliary
weight.

\paragraph{Candidate-pair target.}
For each accepted pair, the frozen teacher target and current student
distribution are renormalized on $\mathcal{S}_i=\{t_i,s_i\}$:
\begin{equation}
\bar p_i^T(v)=\frac{p_i^T(v)}{\sum_{u\in\mathcal{S}_i}p_i^T(u)},
\qquad
\bar q_i^\theta(v)=
\frac{\pi_\theta(v\mid h_S,z_{<k_i})}
     {\sum_{u\in\mathcal{S}_i}\pi_\theta(u\mid h_S,z_{<k_i})},
\qquad v\in\mathcal{S}_i.
\label{eq:pair_normalization}
\end{equation}
The per-position loss is the Jensen-Shannon divergence between these
distributions:
\begin{equation}
\mathcal{L}_{\mathrm{pair}}(i)
=\frac{1}{2}\operatorname{KL}(\bar p_i^T\,\|\,\mu_i)
+\frac{1}{2}\operatorname{KL}(\bar q_i^\theta\,\|\,\mu_i),
\qquad
\mu_i=\frac{\bar p_i^T+\bar q_i^\theta}{2}.
\label{eq:pair_jsd}
\end{equation}
Verification and distillation play complementary roles: executed paired
information gain determines whether a candidate pair is retained and how
strongly it is weighted, while the Jensen--Shannon divergence transfers the
verified teacher preference over the two adjudicated tokens.
\paragraph{Joint training objective.}
Let $\mathcal{V}_Q=\{i\in\mathcal{I}_Q:\Delta\mathrm{IG}_i>0\}$ denote the
accepted query positions. The query-level auxiliary loss and the actor
objective are
\begin{equation}
\mathcal{L}_{\mathrm{IGSD}}(Q)
=\frac{1}{|\mathcal{I}_Q|}
\sum_{i\in\mathcal{V}_Q}w_i\mathcal{L}_{\mathrm{pair}}(i),
\qquad
\mathcal{L}_{\mathrm{actor}}
=\mathcal{L}_{\mathrm{GRPO}}
+\lambda\,\mathbb{E}_{Q}[\mathcal{L}_{\mathrm{IGSD}}(Q)].
\label{eq:joint_objective}
\end{equation}
We normalize by all query-token positions, so rejected positions contribute
zero and the auxiliary term is not inflated solely by the number of accepted
disagreements.
Verification is training-only; inference uses the standard search agent
without a teacher, verification branches, or extra retrieval calls.

\section{Experiments}
\label{sec:exp}

We evaluate IGSD along three axes. \textbf{RQ1:} Does verified
hindsight distillation improve trained search policies? \textbf{RQ2:} Does
executed paired information gain align with downstream utility? \textbf{RQ3:}
How much does each component of the decision chain contribute to the trained
policy?

\subsection{Experimental Setup}
\label{sec:setup}

\paragraph{Datasets.}
Following Search-R1~\citep{jin2025searchr1}, we train on the combined training
splits of NQ and HotpotQA, and evaluate on the evaluation sets of
seven benchmarks: NQ, TriviaQA, and PopQA (single-hop), and HotpotQA,
2WikiMultiHopQA, MuSiQue, and Bamboogle (multi-hop). We report exact-match
(EM) accuracy per dataset and the macro-average across the seven benchmarks.

\paragraph{Baselines.}
Inference-only references are Direct Generation, Vanilla RAG, and
Search-o1~\citep{li2025searcho1}; outcome-reward RL includes
Search-R1~\citep{jin2025searchr1}, ReSearch~\citep{chen2025research}, and
AutoRefine~\citep{shi2025searchrefinethinkfacilitating}; process-level supervision includes
IGPO~\citep{wang2025igpo}, GiGPO~\citep{feng2025gigpo}, CRAFT~\citep{meng2026craft},
SD-Search~\citep{ma2026sdsearch}, and LOTAPO~\citep{zhu2026lotapo}. We also
report IGSD w/o Gate
(Section~\ref{sec:ablation-gate}), the same recipe without verification
gating.

\paragraph{Implementation details.}
We use Qwen2.5-3B-Instruct~\citep{qwen2025} as the main policy and also
evaluate IGSD with Qwen2.5-7B-Instruct. All locally trained methods are
implemented with SearchAgent-Zero~\citep{li2026searchagentzero}, which
builds on the veRL training framework~\citep{sheng2024verl} and adds
search-agent-specific components, including the multi-turn search pipeline
and tool interface. Training uses batches of 256 prompts and five rollouts
per prompt, a learning rate of $10^{-6}$, and 400 optimization steps.
All locally trained runs use eight NVIDIA A100 80GB GPUs.
We use E5-base-v2~\citep{wang2022e5} with the December 2018 Wikipedia
corpus~\citep{karpukhin2020dpr} and
retrieve the top three passages per search call, and the policy may take at most
six assistant turns per trajectory. GRPO uses the terminal answer evaluator's
reward,
and evaluation uses greedy decoding with one response per question; all locally
trained methods are evaluated at the final checkpoint. Full hyperparameter
details are provided in Appendix~\ref{app:hyper}.

\FloatBarrier

\subsection{Main Results (RQ1)}
\label{sec:main}

\begin{table}[t]
\centering
\caption{Main results with Qwen2.5-3B-Instruct. Bold and underline indicate
the best and second-best values in each column.}
\label{tab:main}
\scriptsize
\setlength{\tabcolsep}{3pt}
\resizebox{0.84\textwidth}{!}{%
\begin{tabular}{lcccccccc}
\toprule
& \multicolumn{3}{c}{\textbf{Single-Hop}} & \multicolumn{4}{c}{\textbf{Multi-Hop}} & \\
\cmidrule(lr){2-4}\cmidrule(lr){5-8}
Method & NQ & TriviaQA & PopQA & HotpotQA & 2Wiki & MuSiQue & Bamboogle & Avg. \\
\midrule
\rowcolor{black!10}\multicolumn{9}{l}{\emph{Inference-only}} \\
Direct Generation & 0.106 & 0.288 & 0.108 & 0.149 & 0.244 & 0.020 & 0.024 & 0.134 \\
Vanilla RAG & 0.348 & 0.544 & 0.387 & 0.255 & 0.226 & 0.047 & 0.080 & 0.270 \\
Search-o1 & 0.238 & 0.472 & 0.262 & 0.221 & 0.218 & 0.054 & 0.320 & 0.255 \\
\midrule
\rowcolor{black!10}\multicolumn{9}{l}{\emph{Outcome-reward RL}} \\
Search-R1 & 0.439 & 0.591 & 0.423 & 0.388 & 0.404 & \underline{0.185} & 0.328 & 0.394 \\
ReSearch & 0.387 & 0.581 & 0.410 & 0.302 & 0.236 & 0.124 & 0.266 & 0.329 \\
AutoRefine & 0.436 & 0.597 & 0.411 & 0.397 & 0.418 & \textbf{0.187} & 0.352 & 0.400 \\
\midrule
\rowcolor{black!10}\multicolumn{9}{l}{\emph{Process-level supervision}} \\
IGPO & 0.445 & 0.604 & 0.434 & 0.389 & 0.405 & 0.171 & 0.312 & 0.394 \\
GiGPO & 0.452 & 0.605 & 0.442 & 0.389 & 0.387 & 0.139 & 0.280 & 0.385 \\
CRAFT & 0.436 & 0.594 & 0.414 & 0.387 & 0.410 & 0.176 & 0.368 & 0.398 \\
SD-Search & \textbf{0.460} & 0.612 & 0.424 & \underline{0.415} & \underline{0.428} & 0.181 & 0.384 & 0.415 \\
LOTAPO & 0.434 & 0.585 & 0.425 & 0.389 & 0.395 & 0.178 & 0.336 & 0.392 \\
\midrule
\rowcolor{black!10}\multicolumn{9}{l}{\emph{IGSD and controls}} \\
IGSD w/o Gate & 0.455 & \textbf{0.623} & \underline{0.448} & 0.401 & 0.407 & 0.184 & \underline{0.400} & \underline{0.417} \\
\textbf{IGSD} & \underline{0.458} & \underline{0.620} & \textbf{0.461} & \textbf{0.420} & \textbf{0.443} & 0.183 & \textbf{0.408} & \textbf{0.428} \\
\bottomrule
\end{tabular}%
}
\end{table}

\begin{table}[t]
\centering
\caption{Results with Qwen2.5-7B-Instruct. Notation as in Table~\ref{tab:main}.}
\label{tab:main7b}
\scriptsize
\setlength{\tabcolsep}{3pt}
\resizebox{0.84\textwidth}{!}{%
\begin{tabular}{lcccccccc}
\toprule
& \multicolumn{3}{c}{\textbf{Single-Hop}} & \multicolumn{4}{c}{\textbf{Multi-Hop}} & \\
\cmidrule(lr){2-4}\cmidrule(lr){5-8}
Method & NQ & TriviaQA & PopQA & HotpotQA & 2Wiki & MuSiQue & Bamboogle & Avg. \\
\midrule
Direct Generation & 0.134 & 0.408 & 0.140 & 0.183 & 0.250 & 0.031 & 0.120 & 0.181 \\
Vanilla RAG & 0.349 & 0.585 & 0.392 & 0.299 & 0.235 & 0.058 & 0.208 & 0.304 \\
Search-o1 & 0.151 & 0.443 & 0.131 & 0.187 & 0.176 & 0.058 & 0.296 & 0.206 \\
Search-R1 & 0.477 & 0.647 & \textbf{0.480} & 0.452 & 0.435 & 0.202 & 0.424 & 0.445 \\
SD-Search & \underline{0.492} & 0.652 & 0.469 & \underline{0.472} & \underline{0.438} & \underline{0.228} & \underline{0.440} & \underline{0.456} \\
IGSD w/o Gate & 0.489 & \underline{0.655} & 0.472 & 0.465 & 0.436 & 0.216 & \underline{0.440} & 0.453 \\
\textbf{IGSD} & \textbf{0.496} & \textbf{0.656} & \underline{0.476} & \textbf{0.478} & \textbf{0.489} & \textbf{0.232} & \textbf{0.464} & \textbf{0.470} \\
\bottomrule
\end{tabular}%
}
\end{table}

\begin{wrapfigure}{R}{0.36\textwidth}
\vspace{-6pt}
\centering
\includegraphics[width=\linewidth]{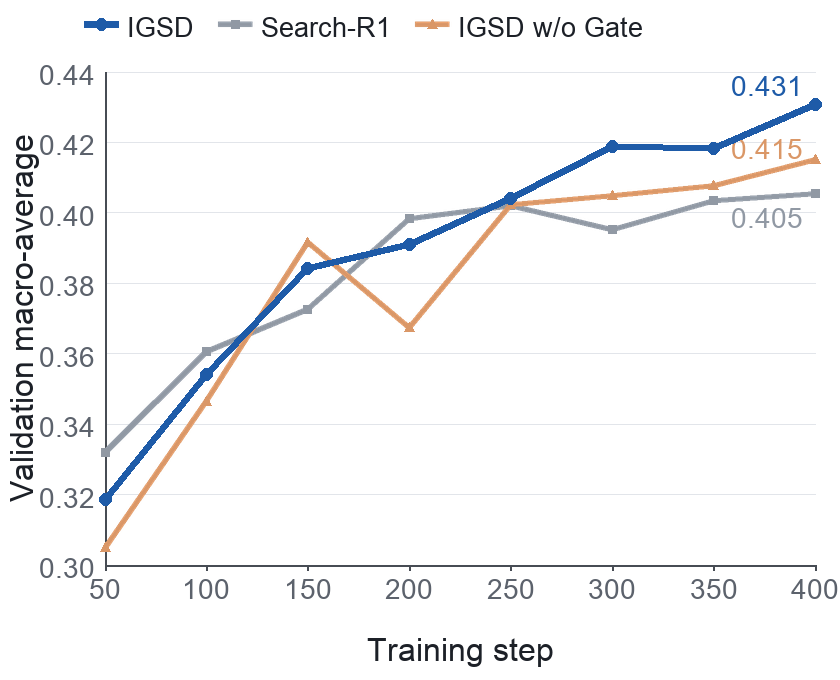}
\caption{Fixed-validation macro-average throughout training for Search-R1,
IGSD w/o Gate, and IGSD.}
\label{fig:validation-dynamics}
\vspace{-8pt}
\end{wrapfigure}

At 3B, IGSD reaches a 0.428 macro-average (Table~\ref{tab:main}). The
results support two conclusions. Against outcome-reward RL, IGSD improves
over the strongest baseline, AutoRefine (0.400), with gains on three of the
four multi-hop benchmarks and a +0.025 gain on 2WikiMultiHopQA,
confirming that step-level supervision adds value beyond trajectory-level
rewards. Against process-level supervision, IGSD exceeds SD-Search (0.415),
with the largest per-dataset separation on PopQA (+0.037), and the ungated
control sharpens the attribution: it shares IGSD's teacher context,
candidate construction, and auxiliary objective, differing only in the
gate, and its 0.417 macro-average therefore assigns the remaining +0.011
to executed-gain weighting. Verified gating is thus a necessary component
of the process signal, not a refinement of it. Three-seed averages preserve
IGSD as the strongest configuration; the complete results are reported in
Appendix~\ref{app:seeds}.

At 7B (Table~\ref{tab:main7b}), IGSD remains ahead, reaching a 0.470
macro-average and retaining a clear advantage over its ungated control (0.453).
It also exceeds Search-R1 (0.445) and SD-Search (0.456), leading on all four
multi-hop benchmarks. The benefit of environment-verified supervision thus
transfers to the larger scale.

Figure~\ref{fig:validation-dynamics} traces the effect of verification on
training stability. Search-R1 improves early but plateaus after roughly the
first 200 steps. IGSD w/o Gate fluctuates throughout training, as unverified
hindsight supervision can help or hurt depending on the current search state.
IGSD continues to improve across the full schedule and shows a smoother
validation trajectory: filtering proposals by their executed gain retains
useful hindsight supervision while suppressing proposals with unfavorable
utility. Complete training-time and search-usage measurements are reported in
Appendix~\ref{app:efficiency}.

\subsection{Mechanistic Evidence (RQ2)}
\label{sec:mechanism}

\begin{figure}[t]
\centering
\includegraphics[width=0.95\textwidth]{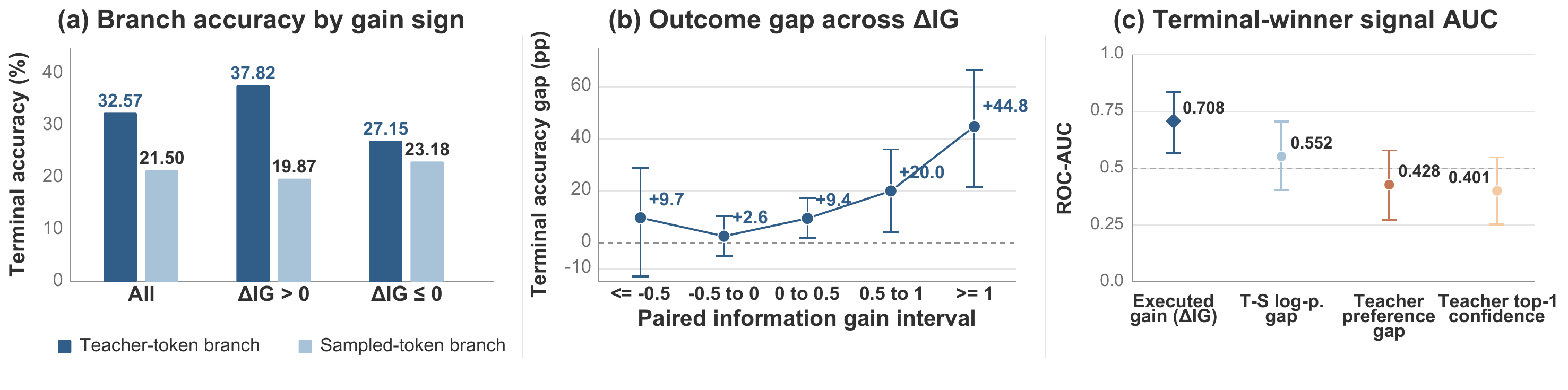}
\caption{Mechanistic evidence for executed paired information gain. Panels
(a)--(c) report branch accuracy by gain sign, teacher--student accuracy gaps
across fixed gain intervals, and terminal-winner ROC-AUC for executed and
model-internal signals. Error bars in (b) and (c) are 95\% question-cluster
bootstrap confidence intervals; both branches use independent completions with
the same frozen student policy and decoding.}
\label{fig:mechanism}
\end{figure}

We next test whether executed paired information gain aligns with downstream
utility. Each candidate pair is evaluated by fixing the candidate tokens in
turn, completing both branches independently with the same frozen student
policy and decoding procedure, and measuring the exact-match (EM) accuracy of
the final answer. Across the 307 evaluated candidate pairs, the teacher-token
branch reaches 0.326 EM on average versus 0.215 for the sampled-token branch.
Splitting the pairs by gain sign (Figure~\ref{fig:mechanism}a) reveals a less
uniform pattern: for $\Delta\mathrm{IG}\leq0$, individual pairs exhibit both
positive and negative teacher--student utility gaps, whereas for
$\Delta\mathrm{IG}>0$, the teacher-token branch shows a larger aggregate
advantage (0.378 vs.\ 0.199 EM).

Figure~\ref{fig:mechanism}b examines the same pairs across fixed
$\Delta\mathrm{IG}$ intervals. Its vertical axis reports the paired
terminal-accuracy gap, defined as teacher-token accuracy minus sampled-token
accuracy in percentage points. The gap is small and variable in the
low-gain region, including $\Delta\mathrm{IG}\leq0$, then increases markedly
for positive gains, reaching +20.0 points for the
$0.5$--$1$ interval and +44.8 points for $\Delta\mathrm{IG}\geq1$. The pattern
supports using executed gain as a graded utility contrast. Both branches share
the same student state, frozen
completion policy, decoding procedure, retriever, and counterfactual control;
only the candidate token fixed at the inspected position differs before the
branches are independently completed and executed. The resulting
$\Delta\mathrm{IG}$ is therefore used as an environment-derived weight for
the auxiliary distillation loss.

Finally, Figure~\ref{fig:mechanism}c compares executed paired information gain
with three model-internal token-probability signals for ranking the
terminal-winning branch among the 72 terminal-discordant pairs. The y-axis
reports ROC-AUC; the four x-axis entries are executed paired information gain,
the teacher--student log-probability gap on the sampled token, the teacher's
log-probability difference between its top-1 and the sampled token, and the
teacher's log-probability for its top-1 token. Executed gain achieves the
highest point estimate (AUC 0.708; 95\% question-cluster bootstrap CI
[0.568, 0.836]), compared with 0.552 for the strongest internal signal, the
sampled-token log-probability gap. This result shows that the executed signal
provides the strongest tested alignment with the terminal-winning branch.

\subsection{Ablation Studies (RQ3)}
\label{sec:ablation}

\begin{figure}[t]
\centering
\includegraphics[width=0.95\textwidth]{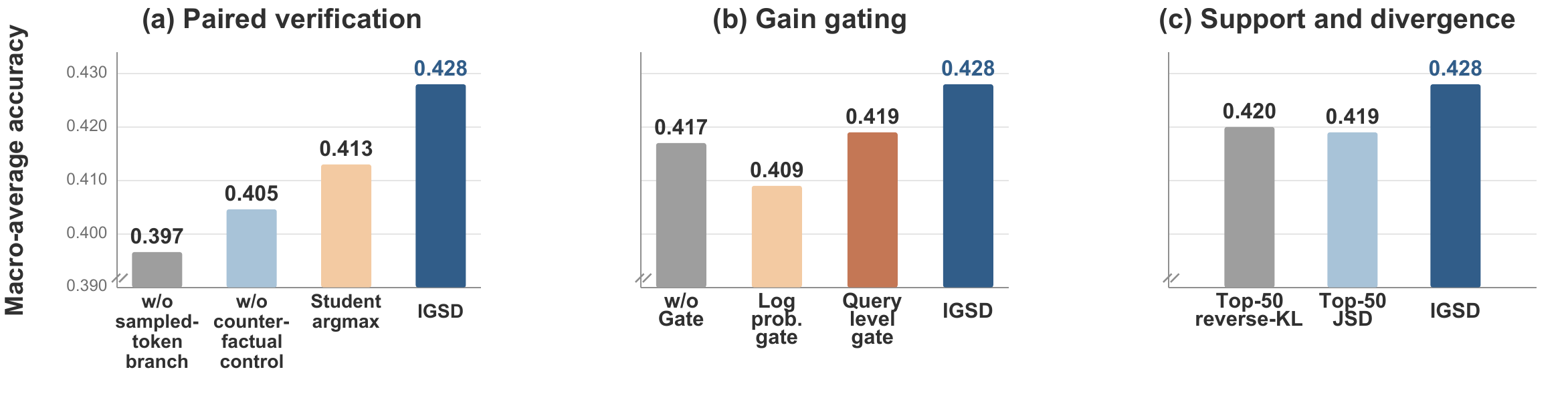}
\caption{Ablation of IGSD's decision chain with the 3B policy. Panels group
variants by paired verification, gain gating, and distillation support and
divergence.}
\label{fig:ablation-summary}
\end{figure}

We next isolate the contributions of IGSD's decision-chain components. Keeping
the default training recipe fixed, we vary supervision scope, paired
verification, gain weighting, and distillation support one at a time.
Table~\ref{tab:mechanism} reports the core comparisons, while
Figure~\ref{fig:ablation-summary} and Appendices~\ref{app:variants}--\ref{app:hyper}
provide the summary and full per-dataset results. Unless explicitly stated,
all variants use the default configuration described in Sections~\ref{sec:method}
and~\ref{sec:setup}.

\begin{table}[H]
\centering
\caption{Component ablations with the Qwen2.5-3B-Instruct policy under the
recipe of Section~\ref{sec:setup}. The best and second-best values per column
are in \textbf{bold} and \underline{underlined}.}
\label{tab:mechanism}
\scriptsize
\setlength{\tabcolsep}{3pt}
\resizebox{0.80\textwidth}{!}{%
\begin{tabular}{lcccccccc}
\toprule
Variant & NQ & TriviaQA & PopQA & HotpotQA & 2Wiki & MuSiQue & Bamboogle & Avg. \\
\midrule
Search-R1 & 0.439 & 0.591 & 0.423 & 0.388 & 0.404 & 0.185 & 0.328 & 0.394 \\
\midrule
All-turn verification & \underline{0.455} & 0.615 & \underline{0.451} & \underline{0.417} & \textbf{0.444} & \textbf{0.201} & 0.368 & \underline{0.421} \\
w/o sampled-token branch & 0.437 & 0.599 & 0.412 & 0.397 & 0.407 & 0.156 & 0.368 & 0.397 \\
w/o counterfactual control & 0.438 & 0.597 & 0.408 & 0.406 & 0.420 & 0.180 & 0.384 & 0.405 \\
w/o Gate & \underline{0.455} & \textbf{0.623} & 0.448 & 0.401 & 0.407 & 0.184 & \underline{0.400} & 0.417 \\
Log-probability gate & 0.450 & 0.609 & 0.438 & 0.412 & 0.421 & \underline{0.186} & 0.344 & 0.409 \\
\midrule
\textbf{IGSD} & \textbf{0.458} & \underline{0.620} & \textbf{0.461} & \textbf{0.420} & \underline{0.443} & 0.183 & \textbf{0.408} & \textbf{0.428} \\
\bottomrule
\end{tabular}%
}
\end{table}

\paragraph{Supervision scope.}
\label{sec:ablation-scope}

Verification supervises the last valid search turn of a failed trajectory.
Verifying every valid turn instead (\emph{all-turn verification}) reaches a
macro-average of 0.421, with no improvement over the 0.428 reference. We
therefore focus on the last valid turn. By then,
the accumulated context makes teacher--student disagreement more likely to
expose a concrete query-level alternative, yielding more targeted candidate
pairs for distillation.

\paragraph{Paired verification.}
\label{sec:ablation-paired}

The paired-verification ablation isolates the two components of the executed
comparison. Removing the sampled-token branch while retaining the other
verification machinery (\emph{w/o sampled-token branch}) gives a macro-average
of 0.397, while removing the counterfactual control (\emph{w/o counterfactual
control}) gives 0.405. Replacing the sampled student token with the student
policy's arg-max token at the same decision position (\emph{student argmax})
gives 0.413. The full paired construction reaches 0.428
(Figure~\ref{fig:ablation-summary}a). Each component removal lowers the
macro-average, supporting the matched sampled-token reference and the shared
counterfactual control in the executed comparison.

\paragraph{Gain gating.}
\label{sec:ablation-gate}

The gate ablation keeps the candidate pairs, teacher target, and loss fixed
while changing only the weight signal. Unit weights without verification
(\emph{w/o Gate}) reach 0.417. Replacing executed gain with the sampled-
token log-probability gap $\log p_T(s_i)-\log p_S(s_i)$ (\emph{log-probability
gate}) lowers the result to 0.409. In the query-level variant, the executed
gains of all verified token positions in one query are averaged and the same
weight is applied to that query's pairs; this reaches 0.419. The per-pair
executed-gain gate reaches 0.428. The gap between the per-pair and query-level
variants indicates that averaging can wash out local differences between
candidate actions, while the lower log-probability-gate result shows that a
model-internal preference is not an adequate substitute for executed utility
(Figure~\ref{fig:ablation-summary}b).

\paragraph{Support and divergence.}
\label{sec:ablation-support}

The final ablation varies the support and divergence of the auxiliary loss.
Widening the two-token target to the de-duplicated teacher--student top-50
union (\emph{top-50 JSD}) gives 0.419, while \emph{top-50 reverse-KL} gives 0.420,
versus 0.428 for candidate-pair JSD. In these variants, broader supports
lower the macro-average, consistent with assigning distillation mass to tokens
that were not directly adjudicated by environment execution
(Figure~\ref{fig:ablation-summary}c; Table~\ref{tab:variants}).

Taken together, the ablations favor a localized decision chain: the teacher
proposal is compared against the action actually sampled by the student, the
comparison is made after query execution, and the resulting paired utility
contrast is applied only to the corresponding candidate pair. Removing one of
these localization steps weakens the final result in the tested setting. The
complete per-dataset variants and counterfactual-draw analysis are deferred to
Appendix~\ref{app:variants}.

\subsection{Qualitative Cases}
\label{sec:cases}

Table~\ref{tab:cases} reports terminal outcomes from the actual teacher-token
and sampled-token branches after they disagree at the inspected decision. Each
branch fixes its candidate token, independently completes the query under the
same frozen student policy, executes retrieval, and follows the resulting
continuation to a terminal answer. In Case A, the student query leaves the
already identified entity unresolved by using the vague phrase \emph{the
actress}. The hindsight teacher instead recognizes Berlinda Tolbert in the
retrieved evidence, produces a query grounded on that entity, and retrieves
the correct sitcom, The Jeffersons ($\Delta\mathrm{IG}=+3.58$, retained). In
Case B, the current evidence has already identified Sigmund Freud. The
sampled-token branch directly queries for Freud's father and retrieves Jacob
Freud, whereas the hindsight teacher redundantly re-encodes the already
resolved relation as a nested query and returns Sigmund Freud again. The
negative $\Delta\mathrm{IG}$ therefore rejects this redundant teacher proposal
($\Delta\mathrm{IG}=-0.198$), rather than distilling it. Further cases are
given in Appendix~\ref{app:cases}.

\begin{table}[H]
\centering
\setlength{\abovecaptionskip}{2pt}
\setlength{\belowcaptionskip}{2pt}
\caption{Representative paired search decisions, one retained and one
rejected by the gate. Blue and orange mark the student and teacher query
spans; $\Delta\mathrm{IG}$ is teacher minus student information gain.}
\label{tab:cases}
\begingroup
\fontsize{6.2}{6.9}\selectfont
\setlength{\tabcolsep}{2pt}
\renewcommand{\arraystretch}{0.88}
\setlength{\emergencystretch}{1em}
\begin{tabularx}{\linewidth}{@{}>{\raggedright\arraybackslash}X@{\hspace{7pt}}>{\raggedright\arraybackslash}X@{}}
\toprule
\rowcolor{igsdCaseGray}[0pt][0pt]
\multicolumn{2}{@{}p{\linewidth}@{}}{%
  \strut\textbf{A\quad Positive $\Delta\mathrm{IG}$ --- hindsight resolves the entity}%
  \hfill\igsdCorrect{RETAIN}\strut}\\
\addlinespace[1pt]
\multicolumn{2}{@{}p{\linewidth}@{}}{%
  \textbf{Question:} An actress in Strange Fruit was better known for a role
  in a CBS sitcom. Which sitcom is this actress better known for than her
  role in Strange Fruit? \textcolor{igsdCaseMuted}{[Reference: The Jeffersons]}}\\
\multicolumn{2}{@{}p{\linewidth}@{}}{%
  \textbf{State:} Strange Fruit is a 2004 film \ldots{} starring Kent Faulcon
  as William Boyals and \textbf{Berlinda Tolbert} as Emma Ayers
  \textcolor{igsdCaseMuted}{(actress identified; her sitcom unresolved).}}\\
\midrule
\textbf{Student-sampled action} & \textbf{Hindsight-teacher proposal}\\
\textbf{Search:} which CBS sitcom was \igsdStudent{the actress}
  better known for than her role in Strange Fruit
&
\textbf{Search:} which CBS sitcom was \igsdTeacher{Berlinda Tolbert}
  better known for than her role in Strange Fruit\\
\textbf{Retrieved:} \textbf{Ricki Lake} \ldots{} had a recurring role on the
  CBS sitcom \textbf{The King of Queens} \ldots{}
&
\textbf{Retrieved:} \textbf{Berlinda Tolbert} \ldots{} best known for her
  role as Jenny Willis Jefferson \ldots{} on the CBS sitcom
  \textbf{The Jeffersons} \ldots{}\\
\textbf{Answer:} \igsdIncorrect{The King of Queens} --- incorrect
& \textbf{Answer:} \igsdCorrect{The Jeffersons} --- correct\\
\multicolumn{2}{@{}p{\linewidth}@{}}{%
  $\Delta\mathrm{IG}=\textcolor{igsdCaseGood}{+3.58>0}$}\\
\midrule
\rowcolor{igsdCaseGray}[0pt][0pt]
\multicolumn{2}{@{}p{\linewidth}@{}}{%
  \strut\textbf{B\quad Negative $\Delta\mathrm{IG}$ --- hindsight repeats the unresolved relation}%
  \hfill\igsdIncorrect{REJECT}\strut}\\
\addlinespace[1pt]
\multicolumn{2}{@{}p{\linewidth}@{}}{%
  \textbf{Question:} Who was the father of the father of psychoanalysis?
  \textcolor{igsdCaseMuted}{[Reference: Jacob Freud]}}\\
\multicolumn{2}{@{}p{\linewidth}@{}}{%
  \textbf{State:} The discipline was established in the early 1890s by
  Austrian neurologist \textbf{Sigmund Freud} and stemmed partly from the
  clinical work of Josef Breuer and others \ldots{}}\\
\midrule
\textbf{Student-sampled action} & \textbf{Hindsight-teacher proposal}\\
\textbf{Search:} who was the father of \igsdStudent{Sigmund Freud}
&
\textbf{Search:} who was the father of the \igsdTeacher{father of psychoanalysis}\\
\textbf{Retrieved:} Jacob Koloman Freud (1815--1896) was the father of
  \textbf{Sigmund Freud}, the founder of
  psychoanalysis. Born in Tysmenytsia in Austrian Galicia \ldots{}
&
\textbf{Retrieved:} The discipline was established in the early 1890s by
  Austrian neurologist
  \textbf{Sigmund Freud} and stemmed partly from the clinical work of Josef
  Breuer and others \ldots{}\\
\textbf{Answer:} \igsdCorrect{Jacob Freud} --- correct
& \textbf{Answer:} \igsdIncorrect{Sigmund Freud} --- incorrect\\
\multicolumn{2}{@{}p{\linewidth}@{}}{%
  $\Delta\mathrm{IG}=\textcolor{igsdCaseBad}{-0.1978<0}$}\\
\bottomrule
\end{tabularx}
\endgroup

\end{table}
\FloatBarrier

\section{Limitations}
\label{sec:limitations}

IGSD relies on reference-answer supervision to estimate retrieved-evidence
utility, so it applies directly to tasks with an explicit answer or reliable
outcome evaluator, but does not directly extend to open-ended tasks lacking an
explicit correctness criterion. Training adds branch completion, execution,
and scoring, but localized verification and a bounded counterfactual budget
keep this cost manageable; inference is unchanged.

\section{Conclusion}
\label{sec:conclusion}

IGSD introduces environment-verified hindsight self-distillation for search
agents. By executing matched teacher and student query proposals from the same
student state and applying shared counterfactual controls, it derives an
executed paired information gain that weights candidate-pair distillation
according to retrieval utility. This grounds auxiliary supervision in
on-policy environment feedback while leaving the GRPO objective and inference
procedure unchanged. Across seven QA benchmarks and both 3B and 7B policies,
IGSD improves macro-average exact-match accuracy over the evaluated baselines,
and executed gain shows stronger alignment with terminal utility than the
tested model-internal signals. These results establish environment-verified
utility as a practical basis for action-level supervision in search agents.

\subsection*{AI Use Statement}

We used generative AI tools to polish the language and presentation of this
manuscript and to support an initial survey and organization of related work.
All technical claims, citations, experimental results, and final wording were
reviewed and verified by the authors, who take full responsibility for the
content of the paper.

\subsection*{Reproducibility statement}

Our code is available at
\url{https://github.com/PhilipGAQ/igsd_code}.
The paper and appendix provide the benchmark statistics, data split
conventions, retrieval and tool settings, prompts, hyperparameters, evaluation
protocol, random-seed results, and ablation configurations used in our study.
These details are intended to make the experimental setup and comparisons
independently checkable.

\bibliography{references}
\bibliographystyle{iclr2027_conference}

\clearpage
\appendix

\section{Experimental Details}
\label{app:details}

\subsection{Data Statistics}
\label{app:data-statistics}

We use the same seven-benchmark collection and split convention as
SD-Search~\citep{ma2026sdsearch}. The datasets are sourced from the FlashRAG
collection. Following Search-R1~\citep{jin2025searchr1}, the training set
combines the training splits of NQ and HotpotQA. We use the test split when it
is available and otherwise use the development split for evaluation.

\begin{table}[H]
\centering
\caption{Dataset statistics for the seven question-answering benchmarks used
in our experiments. A dash indicates that the corresponding split is not
available.}
\label{tab:data-statistics}
\small
\setlength{\tabcolsep}{8pt}
\renewcommand{\arraystretch}{1.08}
\begin{tabular}{lrrrrc}
\toprule
Dataset & Train & Dev & Test & Eval split \\
\midrule
NQ & 79,168 & 8,757 & 3,610 & test \\
TriviaQA & 78,785 & 8,837 & 11,313 & test \\
PopQA & -- & -- & 14,267 & test \\
HotpotQA & 90,447 & 7,405 & -- & dev \\
2WikiMultiHopQA & 15,000 & 12,576 & -- & dev \\
MuSiQue & 19,938 & 2,417 & -- & dev \\
Bamboogle & -- & -- & 125 & test \\
\bottomrule
\end{tabular}
\end{table}

\subsection{Counterfactual Sampling Protocol}
\label{app:counterfactual}

During training, counterfactual document sets are sampled from the batch-level
pool of passages retrieved in the current training batch. Duplicate passages
are removed before sampling. For each candidate
pair, a counterfactual set contains the same number of passages as the
corresponding executed branch; passages retrieved by that branch are excluded,
and the remaining passages are sampled uniformly without replacement. The
same sampled set is used for the teacher and student branches of a pair, and
the default number of shared draws is $C=3$. Counterfactual passages are
serialized with the same tool-response template, passage ordering, and
truncation rule as the corresponding real retrieval response.

\subsection{Verification Coverage}
\label{app:coverage}

Table~\ref{tab:coverage} separates candidate formation from environment
verification. A training instance enters the auxiliary pipeline when a failed
rollout has a successful sibling and a valid last search turn. Within the
supervised query span, only teacher--student disagreements form candidate
pairs; positions where the teacher agrees with the sampled token receive no
paired verification or auxiliary distillation. Candidates with valid matched
executions are then scored by executed paired information gain, and only pairs
with positive gain receive a nonzero auxiliary weight. Rates in the first block
use token-level totals over the 400-step training schedule. The terminal audit
in the second block is a separate diagnostic subset, not an estimate of
candidate prevalence; its sizes are given in the definitions.

\begin{table}[H]
\centering
\caption{Candidate formation, paired verification, and diagnostic-audit
coverage. Training-funnel rates use token-level totals over the 400-step
schedule; the terminal audit is a separate source-stratified subset.}
\label{tab:coverage}
\scriptsize
\setlength{\tabcolsep}{3pt}
\renewcommand{\arraystretch}{1.02}
\resizebox{\textwidth}{!}{%
\begin{tabular}{p{3.15cm}p{9.0cm}>{\raggedleft\arraybackslash}p{2.35cm}}
\toprule
Stage & Definition & Rate \\
\midrule
\multicolumn{3}{l}{\emph{Training-time candidate funnel}} \\
Query-token positions & All supervised query-span positions & 100\% \\
Teacher--student disagreements & $t_i \ne s_i$ & 9.22\% of query tokens \\
Submitted to paired verification & Disagreements sent to matched evaluation & 99.68\% of disagreements \\
Valid paired evaluations & Submitted pairs passing both execution and scoring checks & 99.90\% of submitted \\
Positive-gain pairs & Valid pairs with $\Delta\mathrm{IG}>0$ & 53.45\% of valid pairs \\
Positive-gate query tokens & Positions with nonzero auxiliary weight & 4.91\% of query tokens \\
\midrule
\multicolumn{3}{l}{\emph{Diagnostic terminal audit}} \\
Full diagnostic candidate pool & Candidate pairs before terminal-audit subsampling (4{,}984 pairs) & 100\% of pool \\
Terminal-ready audit pairs & Source-stratified, quota-targeted valid-terminal subset (307 pairs) & 6.16\% of pool \\
Question clusters & Questions represented in the audit (70 clusters) & -- \\
\bottomrule
\end{tabular}
}%
\end{table}

\subsection{Prompts and Tool Interface}
\label{app:prompts}

Table~\ref{tab:prompts} summarizes the prompt components used in our
experiments. The base question prompt, privileged teacher instruction, and
Hermes-style search schema enter the trajectory at different stages. 
The question itself is supplied by the
dataset; it is not rewritten by IGSD.

\begin{table}[H]
\centering
\caption{Prompts and search-tool interface used in our experiments.}
\label{tab:prompts}
\scriptsize
\setlength{\tabcolsep}{3pt}
\renewcommand{\arraystretch}{1.12}
\begin{tabular}{p{2.45cm}p{9.4cm}}
\toprule
Component & Prompt or template \\
\midrule
Base system prompt &
\parbox[t]{9.4cm}{\ttfamily\scriptsize
You are a helpful and harmless assistant. Answer the given question. Reason
between <thought> and </thought> after each new observation. If external
information is needed, emit one JSON object inside <tool\_call> tags. The
search tool returns results inside <tool\_response> tags. Break complex
questions into specific sub-questions, keep queries unique, and finish with
<answer>...</answer> when no further search is needed.} \\
\midrule
Search tool schema &
\parbox[t]{9.4cm}{\ttfamily\scriptsize
Hermes/XML tool call: <tool\_call>\{``name'': ``search'', ``arguments'':
\{``query\_list'': [``fully-formed semantic query'', ...]\}\}</tool\_call>.
The schema exposes one function, \texttt{search}; \texttt{query\_list} is a
non-empty list of strings.} \\
\midrule
Hindsight teacher instruction &
\parbox[t]{9.4cm}{\ttfamily\scriptsize
You are the query-repair branch of a search agent. Propose exactly one next
search action for the failed trajectory prefix below. A successful sibling's
search queries and score are strategy hints only. Do not answer the question,
discuss the hint, or emit thought. Output exactly one <tool\_call> for the
search tool. The prompt then appends: Failed trajectory prefix; Successful
sibling score; Successful sibling search queries.} \\
\bottomrule
\end{tabular}
\end{table}

\section{Signal Diagnostics}
\label{app:diagnostics}

Table~\ref{tab:diagnostics} evaluates every signal against the same labels:
which branch of a terminal-discordant pair ultimately answers correctly. Of
the 307 evaluated pairs, 53 are teacher-only correct, 19 student-only correct,
47 both correct, and 188 both wrong; the 72 discordant pairs, drawn from 70
questions, define the winner labels. Confidence intervals are 95\%
question-cluster percentile bootstrap intervals over 20{,}000 resamples. The
table compares executed gain with model-internal token scores; only
executed gain is used by IGSD. Executed gain has the highest AUC among the
tested signals and, under its zero-threshold sign, selects the winning branch
on 47 of 72 pairs (65.3\%, CI $[54.2\%, 76.1\%]$).

\begin{table}[H]
\centering
\caption{Predicting the paired terminal winner from model-internal
token-probability scores and executed gain.}
\label{tab:diagnostics}
\small
\setlength{\tabcolsep}{2pt}
\begin{tabular}{llc}
\toprule
Signal family & Signal & Terminal-winner AUC (95\% CI) \\
\midrule
Internal confidence & Teacher top-1 log-probability & 0.401 [0.252, 0.548] \\
Internal likelihood & Teacher--student sampled-token log-probability gap & 0.552 [0.404, 0.706] \\
Internal preference & Teacher preference gap & 0.428 [0.272, 0.579] \\
Environment execution & Executed paired information gain & \textbf{0.708 [0.568, 0.836]} \\
\bottomrule
\end{tabular}
\end{table}

\begin{figure}[H]
\centering
\includegraphics[width=0.66\textwidth]{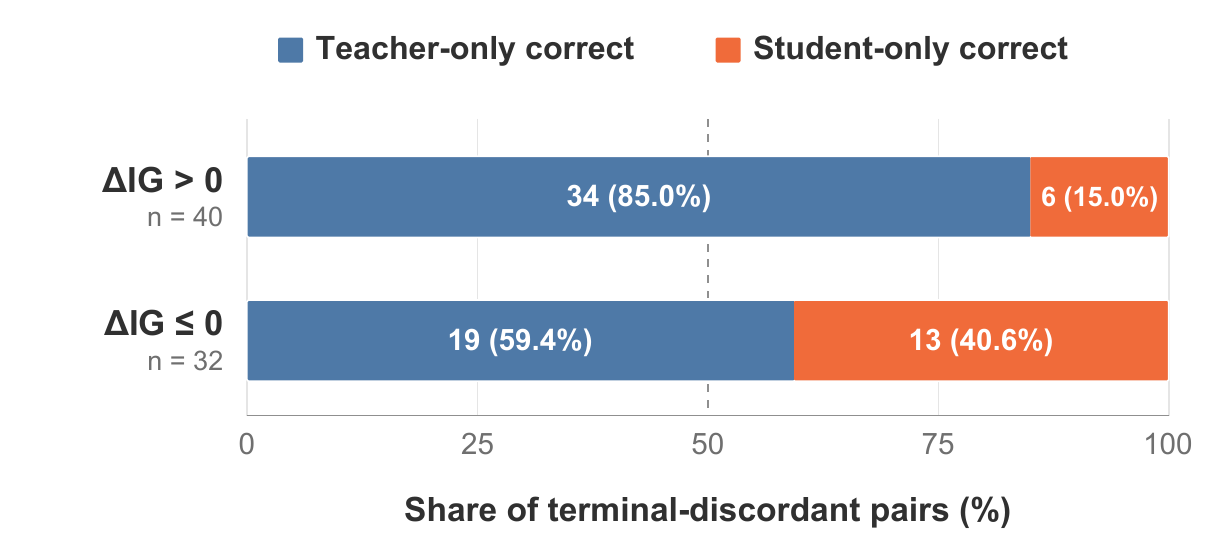}
\caption{Terminal-outcome composition among terminal-discordant matched pairs. Bars
partition pairs by which branch is correct; labels give counts and within-group
percentages. Positive executed gain contains a larger share of
teacher-winning outcomes.}
\label{fig:gain-outcome}
\end{figure}

Figure~\ref{fig:gain-outcome} breaks down the terminal-discordant pairs by which branch is
correct. Positive executed gain contains 34 teacher-only correct
and 6 student-only correct pairs, compared with 19 and 13, respectively, in the
non-positive group. The gate therefore retains a substantially higher share of
teacher-winning outcomes.

\noindent
\begin{minipage}[t]{0.54\textwidth}
\vspace{0pt}
Figure~\ref{fig:utility-agreement} tracks executed gain and three-way action
agreement over training. The executed gain rises early, reaches
a mid-training maximum, and then decreases while remaining positive; agreement
among the teacher top-1 action, student top-1 action, and sampled student action
increases over the same horizon. Together, the curves show that useful teacher
guidance is strongest early in training while the three actions increasingly
coincide later in training.
\end{minipage}\hfill
\begin{minipage}[t]{0.42\textwidth}
\vspace{0pt}
\centering
\includegraphics[width=\linewidth]{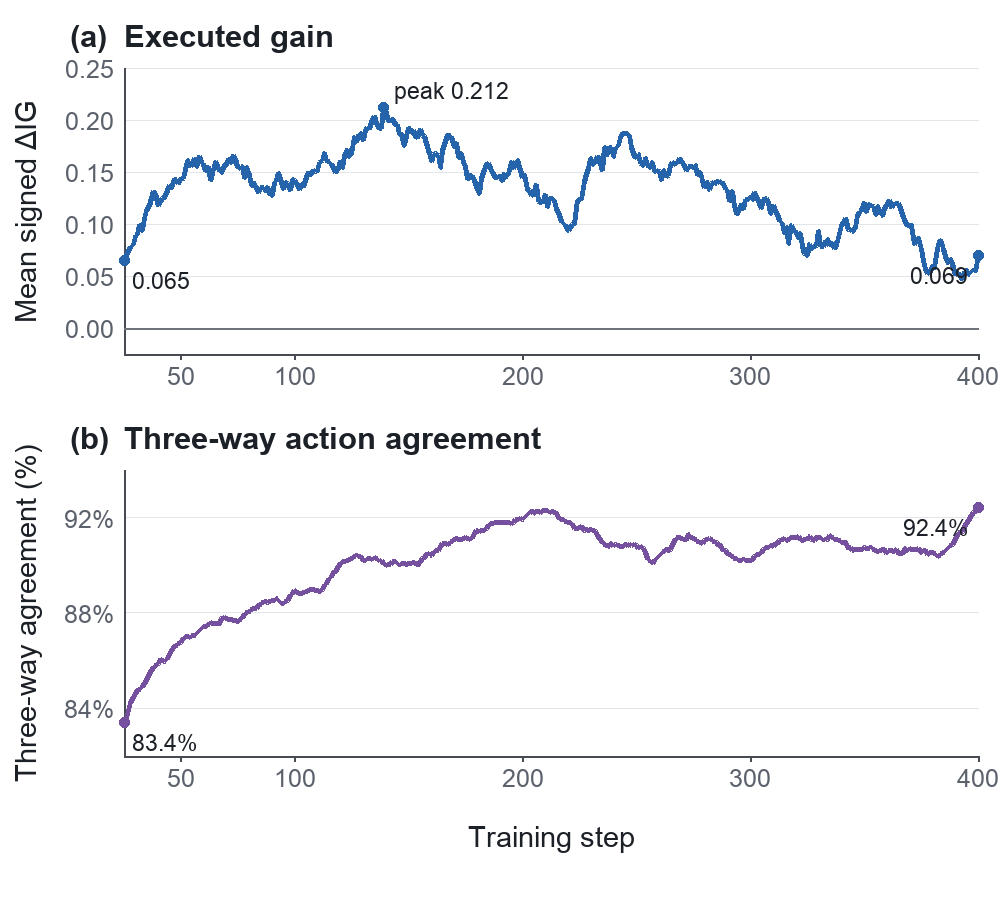}
\refstepcounter{figure}
\par\vspace{2pt}
\raggedright\footnotesize
\textbf{Figure~\thefigure:} Mean signed executed gain and three-way action
agreement over training.
\label{fig:utility-agreement}
\end{minipage}
\par\medskip


\FloatBarrier

\section{Repeated-Run Results}
\label{app:seeds}

Table~\ref{tab:seeds} reports three-seed means $\pm$ standard deviations for
the four core configurations.

\begin{table}[H]
\centering
\caption{Repeated-run results with three random seeds (Qwen2.5-3B-Instruct,
mean $\pm$ standard deviation). The best mean in each column is in bold.}
\label{tab:seeds}
\scriptsize
\setlength{\tabcolsep}{3pt}
\resizebox{\textwidth}{!}{%
\begin{tabular}{lcccccccc}
\toprule
Method & NQ & TriviaQA & PopQA & HotpotQA & 2Wiki & MuSiQue & Bamboogle & Avg. \\
\midrule
Search-R1 & 0.438 $\pm$ 0.009 & 0.587 $\pm$ 0.006 & 0.429 $\pm$ 0.010 & 0.382 $\pm$ 0.010 & 0.420 $\pm$ 0.012 & 0.190 $\pm$ 0.011 & 0.340 $\pm$ 0.027 & 0.398 $\pm$ 0.006 \\
SD-Search & 0.455 $\pm$ 0.010 & 0.615 $\pm$ 0.006 & 0.427 $\pm$ 0.008 & 0.418 $\pm$ 0.015 & 0.419 $\pm$ 0.010 & 0.178 $\pm$ 0.008 & 0.400 $\pm$ 0.022 & 0.416 $\pm$ 0.008 \\
IGSD w/o Gate & 0.450 $\pm$ 0.008 & \textbf{0.621} $\pm$ 0.009 & 0.450 $\pm$ 0.012 & 0.402 $\pm$ 0.015 & 0.400 $\pm$ 0.009 & \textbf{0.191} $\pm$ 0.013 & 0.384 $\pm$ 0.026 & 0.414 $\pm$ 0.008 \\
\textbf{IGSD} & \textbf{0.460} $\pm$ 0.008 & 0.618 $\pm$ 0.010 & \textbf{0.459} $\pm$ 0.010 & \textbf{0.421} $\pm$ 0.012 & \textbf{0.439} $\pm$ 0.013 & 0.190 $\pm$ 0.009 & \textbf{0.404} $\pm$ 0.022 & \textbf{0.427} $\pm$ 0.006 \\
\bottomrule
\end{tabular}%
}
\end{table}

\section{Efficiency and Search Usage}
\label{app:efficiency}

Table~\ref{tab:efficiency} separates the measured 400-step training cost from
inference-time tool use; both are wall-clock measurements under the
corresponding experimental settings. Table~\ref{tab:search-calls} gives the
dataset-level breakdown of evaluation-time search calls for the same three
methods (Search-R1, SD-Search, and IGSD).

\begin{table}[H]
\centering
\caption{Training and inference efficiency over 400 optimization steps.
Search-call columns report inference-time tool use on the same 51,713-example
greedy evaluation set; wall-clock values are hours:minutes:seconds and
calls/example is reported to three decimals.}
\label{tab:efficiency}
\small
\begin{tabular}{lcccc}
\toprule
Method & Wall-clock time (400 steps) & Relative time & Search calls & Calls/example \\
\midrule
Search-R1 & 23:41:12 & 1.00$\times$ & 84,393 & 1.632 \\
SD-Search & 51:35:04 & 2.18$\times$ & 105,629 & 2.043 \\
IGSD & 29:42:53 & 1.25$\times$ & 85,366 & 1.651 \\
\bottomrule
\end{tabular}
\end{table}

\begin{table}[H]
\centering
\caption{Average evaluation-time search calls per example by dataset.}
\label{tab:search-calls}
\small
\begin{tabular}{lccc}
\toprule
Dataset & Search-R1 & IGSD & SD-Search \\
\midrule
NQ & 1.1626 & 1.1573 & 2.0003 \\
TriviaQA & 1.2766 & 1.2547 & 2.0069 \\
PopQA & 1.2267 & 1.2931 & 1.9990 \\
HotpotQA & 1.8913 & 1.8967 & 2.0330 \\
2WikiMultiHopQA & 2.2491 & 2.2609 & 2.1036 \\
MuSiQue & 2.3645 & 2.4075 & 2.2420 \\
Bamboogle & 1.9840 & 1.9920 & 2.0480 \\
Overall & 1.6319 & 1.6508 & 2.0426 \\
\bottomrule
\end{tabular}
\end{table}

\section{Extended Ablations}
\label{app:variants}

\subsection{Hindsight Context Ablation}
\label{app:hindsight}

\begin{figure}[H]
\centering
\includegraphics[width=\textwidth]{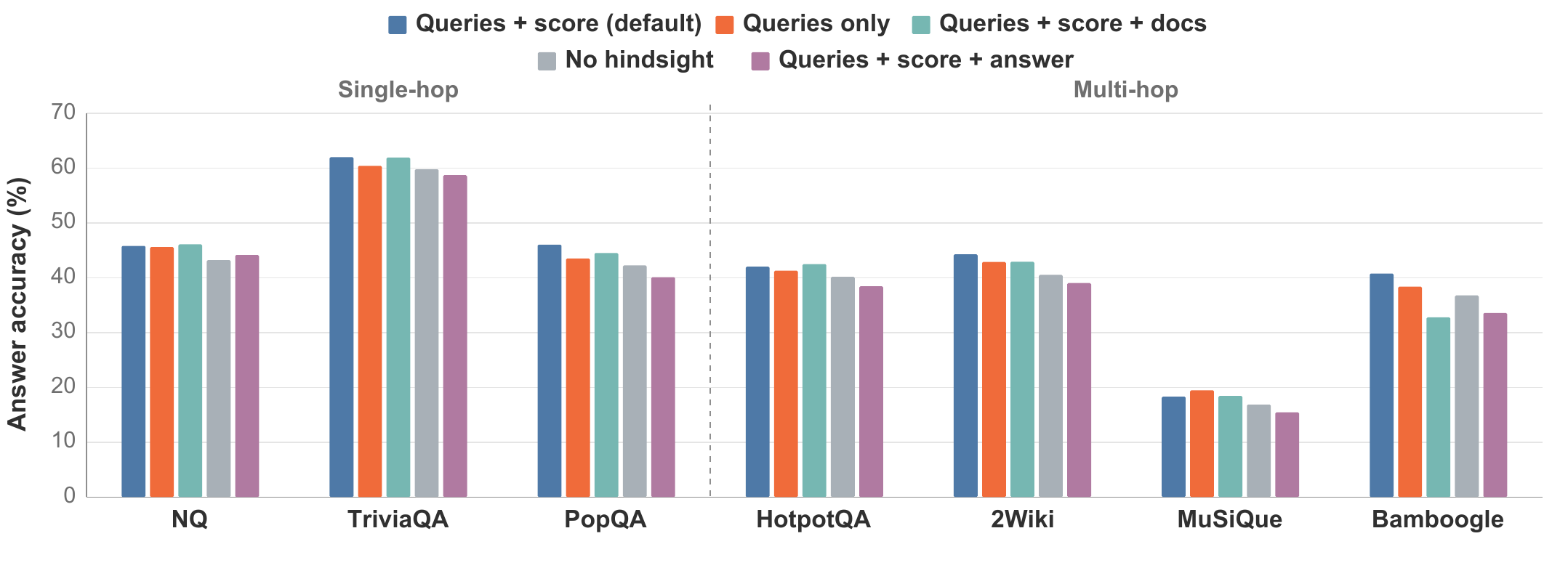}
\caption{Complete hindsight-context comparison. The default teacher context is
successful sibling queries plus outcome score; other bars add, remove, or
replace this privileged information as described in the text.}
\label{fig:hindsight}
\end{figure}

Figure~\ref{fig:hindsight} gives the complete per-dataset comparison of
hindsight contexts. The variants change only the privileged information
supplied to the self-teacher. The default ``Queries + score'' context is the
strongest setting (macro-average $0.428$). Removing the sibling outcome score
(``Queries only'') lowers the average to $0.417$, adding retrieved documents
(``Queries + score + docs'') reaches $0.413$, and removing hindsight entirely
(``No hindsight'') reaches $0.400$. Supplying the ground-truth answer
(``Queries + score + answer'') gives the lowest average, $0.385$. This is an
intentionally leaky context: exposing the answer can encourage answer-conditioned
behavior and reward hacking rather than useful retrieval, so we include it as a
stress test of privileged information.

\subsection{Complete Variant Comparisons}

Table~\ref{tab:variants} collects the complete per-dataset results for the
variants discussed in Section~\ref{sec:ablation}: its reference and ablation rows
are those of Table~\ref{tab:mechanism}, and the remaining blocks add the
support and draw-count variants. The first block is the IGSD reference;
each subsequent block varies one stage of the method. All rows use the 3B
policy and the training recipe of Section~\ref{sec:setup}.

The supervision-scope block verifies every valid search turn instead of only the
last one, with the teacher hint, the gate, and the loss unchanged. The next block
varies how a teacher proposal is verified. IGSD compares
the teacher's top-1 token with the student's sampled token under a per-candidate
matched completion; the w/o sampled-token-branch row removes the matched
sampled-token reference, the w/o counterfactual-control row removes the shared
counterfactual control, and the student-argmax row replaces the sampled
student token with the student's arg-max token at the same position. The gate
block varies how the executed gain weights
supervision: \emph{log-probability gate} substitutes the sampled-token
log-probability gap $\log p_T(s_i)-\log p_S(s_i)$ for $\Delta\mathrm{IG}_i$,
the binary-gate row replaces the rectified-tanh gate with a sign gate,
\emph{query-level gate} averages the gain over a query's verified pairs, and
\emph{w/o Gate} issues no verification branches. The support-and-divergence block
widens the two-token target to the de-duplicated union of the teacher and
student top-50 tokens and to a top-50 reverse-KL objective. The final block
varies the number of counterfactual draws.

\begin{table}[H]
\centering
\caption{Complete per-dataset comparisons of the supervision-scope, verification,
gating, support, and counterfactual-draw variants with the 3B policy. Best value
in each column is in bold.}
\label{tab:variants}
\scriptsize
\setlength{\tabcolsep}{2.5pt}
\resizebox{0.94\textwidth}{!}{%
\begin{tabular}{lcccccccc}
\toprule
Variant & NQ & Triv. & Pop. & Hot. & 2Wiki & Mus. & Bam. & Avg. \\
\midrule
\rowcolor{black!8}\multicolumn{9}{l}{\emph{Reference}} \\
\textbf{IGSD (default)} & 0.458 & 0.620 & \textbf{0.461} & 0.420 & 0.443 & 0.183 & \textbf{0.408} & \textbf{0.428} \\
\midrule
\rowcolor{black!8}\multicolumn{9}{l}{\emph{Supervision scope}} \\
All-turn verification & 0.455 & 0.615 & 0.451 & 0.417 & 0.444 & 0.201 & 0.368 & 0.421 \\
\midrule
\rowcolor{black!8}\multicolumn{9}{l}{\emph{Paired verification}} \\
Student argmax & 0.456 & 0.618 & 0.443 & 0.409 & 0.425 & 0.179 & 0.360 & 0.413 \\
w/o sampled-token branch & 0.437 & 0.599 & 0.412 & 0.397 & 0.407 & 0.156 & 0.368 & 0.397 \\
w/o counterfactual control & 0.438 & 0.597 & 0.408 & 0.406 & 0.420 & 0.180 & 0.384 & 0.405 \\
\midrule
\rowcolor{black!8}\multicolumn{9}{l}{\emph{Gate}} \\
Log-probability gate & 0.450 & 0.609 & 0.438 & 0.412 & 0.421 & 0.186 & 0.344 & 0.409 \\
Binary gate & 0.456 & 0.617 & 0.444 & 0.419 & 0.439 & 0.197 & 0.376 & 0.421 \\
Query-level gate & \textbf{0.469} & 0.617 & 0.430 & 0.421 & 0.434 & 0.175 & 0.384 & 0.419 \\
w/o Gate & 0.455 & \textbf{0.623} & 0.448 & 0.401 & 0.407 & 0.184 & 0.400 & 0.417 \\
\midrule
\rowcolor{black!8}\multicolumn{9}{l}{\emph{Support and divergence}} \\
Top-50 JSD & 0.455 & 0.611 & 0.441 & \textbf{0.424} & \textbf{0.445} & 0.188 & 0.368 & 0.419 \\
Top-50 reverse-KL & 0.457 & 0.613 & 0.433 & 0.414 & 0.428 & \textbf{0.204} & 0.392 & 0.420 \\
\midrule
\rowcolor{black!8}\multicolumn{9}{l}{\emph{Counterfactual draws}} \\
$C=1$ counterfactual draw & 0.460 & 0.614 & 0.439 & 0.419 & 0.432 & 0.187 & 0.400 & 0.422 \\
$C=5$ counterfactual draws & 0.449 & 0.609 & 0.422 & 0.416 & 0.416 & 0.183 & 0.384 & 0.411 \\
\bottomrule
\end{tabular}%
}
\end{table}

\section{Hyperparameter Sensitivity}
\label{app:hyper}

\begin{figure}[H]
\centering
\includegraphics[width=0.70\textwidth]{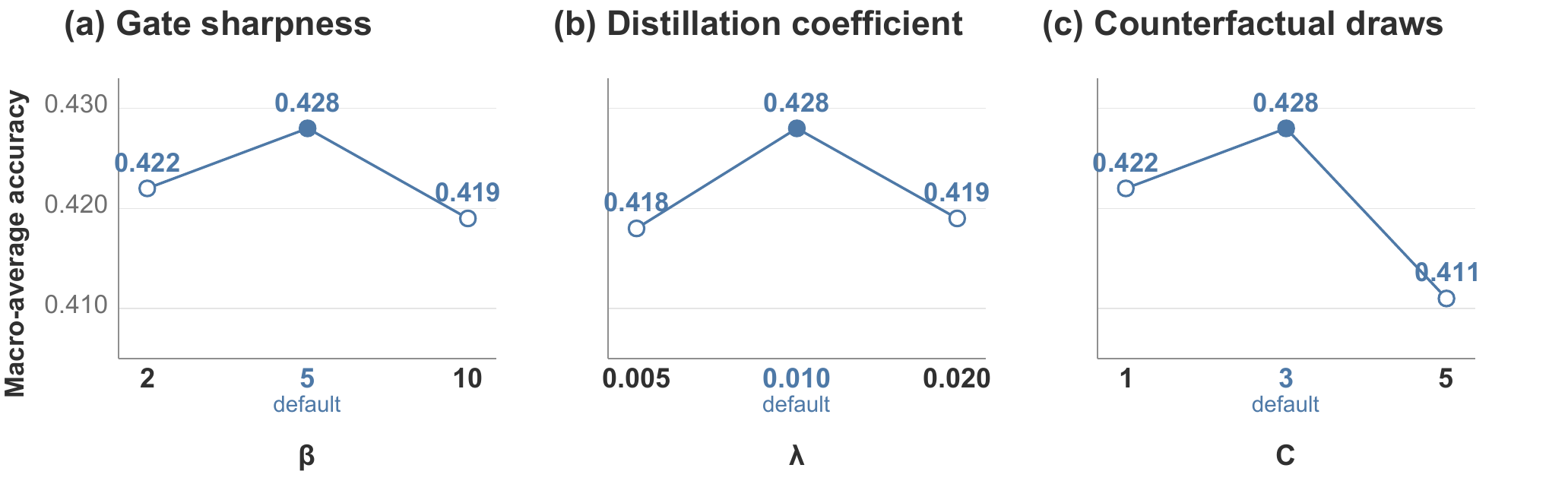}
\caption{Hyperparameter sensitivity of IGSD with the 3B policy. Gate sharpness,
distillation coefficient, and counterfactual draws are varied one at
a time.}
\label{fig:hyper}
\end{figure}

Figure~\ref{fig:hyper} varies one hyperparameter at a time around the default
configuration. The default reaches the highest macro-average in each sweep.

Table~\ref{tab:hyperparameters} summarizes the reproducibility-critical
settings for the main seven-benchmark experiments. Following common reporting
practice, we include the optimization, retrieval, and method-specific
parameters that affect the training objective or agent behavior.

\begin{table}[H]
\centering
\caption{Training, retrieval, and IGSD hyperparameters.}
\label{tab:hyperparameters}
\scriptsize
\setlength{\tabcolsep}{5pt}
\renewcommand{\arraystretch}{1.05}
\begin{tabular}{p{4.5cm}p{6.8cm}} 
\toprule
Setting & Main experiments \\
\midrule
\multicolumn{2}{l}{\emph{Training configuration}} \\
Policy & Qwen2.5-3B-Instruct; Qwen2.5-7B-Instruct \\
Hardware & 8$\times$ NVIDIA A100 80GB \\
Training data & NQ + HotpotQA \\
Training batch size & 256 \\
Rollouts per prompt & 5 \\
Actor learning rate & $10^{-6}$ \\
Optimization steps & 400 \\
Maximum prompt / response length & 4096 / 3000 \\
Maximum assistant turns & 6 \\
Rollout temperature / top-$p$ & 1.0 / 1.0 \\
\midrule
\multicolumn{2}{l}{\emph{Retrieval configuration}} \\
Retriever & E5-base-v2 \\
Corpus & Wikipedia (December 2018) \\
Passages per query & 3 \\
\midrule
\multicolumn{2}{l}{\emph{IGSD-specific configuration}} \\
Teacher context & Successful sibling queries + outcome score \\
Supervised turn & Last valid search turn \\
Distillation objective & Candidate-pair JSD \\
Distillation span & Search-query tokens \\
Counterfactual draws $C$ & 3 \\
Gate sharpness $\beta$ & 5 \\
Distillation coefficient $\lambda$ & 0.01 \\
\bottomrule
\end{tabular}
\end{table}

\section{Additional Case Summaries}
\label{app:cases}

Table~\ref{tab:cases-all} gives three additional positive-gain paired
trajectories and one negative-gain example with independently completed
terminal answers. Each case retains the question and state, the sampled-token
and teacher-token search branches, retrieved snippets, and the resulting
$\Delta\mathrm{IG}$. Blue and orange mark the local student and teacher query
choices, while gray text explains their retrieval consequences.

\begin{table}[H]
\centering
\caption{Additional paired trajectories. The table includes three positive-gain
cases and one negative-gain case; both branches are completed independently
with the same frozen student policy and decoding procedure.}
\label{tab:cases-all}
\begingroup
\fontsize{6.8}{7.55}\selectfont
\setlength{\tabcolsep}{2.5pt}
\renewcommand{\arraystretch}{0.98}
\setlength{\emergencystretch}{1em}
\begin{tabularx}{\linewidth}{@{}>{\raggedright\arraybackslash}X@{\hspace{7pt}}>{\raggedright\arraybackslash}X@{}}
\toprule
\rowcolor{igsdCaseGray}[0pt][0pt]
\multicolumn{2}{@{}p{\linewidth}@{}}{%
  \strut\textbf{Positive $\Delta\mathrm{IG}$ --- advances to the missing organizational hop}%
  \hfill\igsdCorrect{RETAIN}\strut}\\
\addlinespace[1pt]
\multicolumn{2}{@{}p{\linewidth}@{}}{%
  \textbf{Question:} What company is Frederic Ewen's employer part of?
  \textcolor{igsdCaseMuted}{[Reference: City University of New York]}}\\
\multicolumn{2}{@{}p{\linewidth}@{}}{%
  \textbf{State:} A previous search asks who Frederic Ewen and his employer are;
  the retrieved evidence identifies him as an English professor at
  \textbf{Brooklyn College} \textcolor{igsdCaseMuted}{(employer resolved; parent
  organization unresolved).}}\\
\midrule
\textbf{Student-sampled action} & \textbf{Hindsight-teacher proposal}\\
\textbf{Search:} what is the \igsdStudent{\texttt{employer of Frederic Ewen}}
&
\textbf{Search:} what is \igsdTeacher{\texttt{Brooklyn College}}\\
\textbf{Retrieved:} Frederic Ewen (1899--1988) was an English professor at
Brooklyn College from 1930 to 1952. During the McCarthy period Ewen was forced
to resign his teaching position \ldots{}
&
\textbf{Retrieved:} Brooklyn College is a university of the
\textbf{City University of New York}, located in the Flatbush neighborhood of
Brooklyn, New York City \ldots{}\\
\textbf{Continuation:} ``\ldots{} his employer was Brooklyn College.''
\newline\textcolor{igsdCaseMuted}{\emph{Repeats the solved employer lookup and
returns the intermediate entity.}}
&
\textbf{Continuation:} ``Brooklyn College is a part of the City University of
New York. Therefore Frederic Ewen's employer is part of it.''
\newline\textcolor{igsdCaseMuted}{\emph{Queries the resolved entity and advances
to the missing hop.}}\\
\textbf{Answer:} \igsdIncorrect{Brooklyn College} --- incorrect
&
\textbf{Answer:} \igsdCorrect{City University of New York} --- correct\\
\multicolumn{2}{@{}p{\linewidth}@{}}{%
  $\Delta\mathrm{IG}=\textcolor{igsdCaseGood}{+3.7049>0}$}\\
\midrule
\rowcolor{igsdCaseGray}[0pt][0pt]
\multicolumn{2}{@{}p{\linewidth}@{}}{%
  \strut\textbf{Positive $\Delta\mathrm{IG}$ --- grounds the target country}%
  \hfill\igsdCorrect{RETAIN}\strut}\\
\addlinespace[1pt]
\multicolumn{2}{@{}p{\linewidth}@{}}{%
  \textbf{Question:} Who is the president of the newly declared independent
  country in the Commission of Truth and Friendship with the country where
  Gado-gado is found? \textcolor{igsdCaseMuted}{[Reference: Francisco Guterres]}}\\
\multicolumn{2}{@{}p{\linewidth}@{}}{%
  \textbf{State:} Retrieved evidence identifies Gado-gado as Indonesian and the
  commission as an Indonesia--East Timor body
  \textcolor{igsdCaseMuted}{(target country resolved; president unresolved).}}\\
\midrule
\textbf{Student-sampled action} & \textbf{Hindsight-teacher proposal}\\
\textbf{Search:} who is the president of the newly declared independent
country in the commission with \igsdStudent{\texttt{Indonesia}}
&
\textbf{Search:} who is the president of \igsdTeacher{\texttt{East Timor}}\\
\textbf{Retrieved:} The Indonesia--Timor Leste Commission on Truth and
Friendship was established jointly by the governments of Indonesia and East
Timor \ldots{}
&
\textbf{Retrieved:} \textbf{Francisco Guterres} is an East Timorese politician
who has been President of East Timor since 2017 \ldots{}\\
\textbf{Continuation:} ``\ldots{} the president of Indonesia is Susilo Bambang
Yudhoyono.''
\newline\textcolor{igsdCaseMuted}{\emph{The long relation leaves Indonesia as
the dominant but incorrect anchor.}}
&
\textbf{Continuation:} ``\ldots{} the president of East Timor is Francisco
Guterres, and East Timor is the country in the commission with Indonesia.''
\newline\textcolor{igsdCaseMuted}{\emph{Names the resolved country and removes
the competing Indonesia anchor.}}\\
\textbf{Answer:} \igsdIncorrect{Susilo Bambang Yudhoyono} --- incorrect
&
\textbf{Answer:} \igsdCorrect{Francisco Guterres} --- correct\\
\multicolumn{2}{@{}p{\linewidth}@{}}{%
  $\Delta\mathrm{IG}=\textcolor{igsdCaseGood}{+3.3852>0}$}\\
\addlinespace[4pt]
\midrule
\rowcolor{igsdCaseGray}[0pt][0pt]
\multicolumn{2}{@{}p{\linewidth}@{}}{%
  \strut\textbf{Positive $\Delta\mathrm{IG}$ --- carries forward the resolved director}%
  \hfill\igsdCorrect{RETAIN}\strut}\\
\addlinespace[1pt]
\multicolumn{2}{@{}p{\linewidth}@{}}{%
  \textbf{Question:} Who is the spouse of the director of California (1947)?
  \textcolor{igsdCaseMuted}{[Reference: Maureen O'Sullivan]}}\\
\multicolumn{2}{@{}p{\linewidth}@{}}{%
  \textbf{State:} Retrieved evidence identifies \textbf{John Farrow} as the
  director \textcolor{igsdCaseMuted}{(director resolved; spouse unresolved).}}\\
\midrule
\textbf{Student-sampled action} & \textbf{Hindsight-teacher proposal}\\
\textbf{Search:} who was the spouse of the director of
\igsdStudent{\texttt{California (1947)}}
&
\textbf{Search:} who was the spouse of \igsdTeacher{\texttt{John Farrow}}\\
\textbf{Retrieved:} California is a 1947 American western film directed by
John Farrow and featuring Ray Milland, Barbara Stanwyck, and \ldots{}
&
\textbf{Retrieved:} Prudence Farrow married teacher Albert Morrill Bruns in
December 1969. She is the daughter of film director John Farrow and actress
Maureen O'Sullivan \ldots{}\\
\textbf{Continuation:} ``\ldots{} the director was John Farrow. John Farrow was
married to Marsha Hunt.''
\newline\textcolor{igsdCaseMuted}{\emph{Keeps the film as the anchor and mistakes
an associated actress for the spouse.}}
&
\textbf{Continuation:} ``\ldots{} the spouse of John Farrow was Maureen
O'Sullivan.''
\newline\textcolor{igsdCaseMuted}{\emph{Carries the resolved director into the
query and retrieves the spouse relation directly.}}\\
\textbf{Answer:} \igsdIncorrect{Marsha Hunt} --- incorrect
&
\textbf{Answer:} \igsdCorrect{Maureen O'Sullivan} --- correct\\
\multicolumn{2}{@{}p{\linewidth}@{}}{%
  $\Delta\mathrm{IG}=\textcolor{igsdCaseGood}{+2.6150>0}$}\\
\addlinespace[4pt]
\midrule
\rowcolor{igsdCaseGray}[0pt][0pt]
\multicolumn{2}{@{}p{\linewidth}@{}}{%
  \strut\textbf{Negative $\Delta\mathrm{IG}$ --- anchors on the wrong newspaper format}%
  \hfill\igsdIncorrect{REJECT}\strut}\\
\addlinespace[1pt]
\multicolumn{2}{@{}p{\linewidth}@{}}{%
  \textbf{Question:} Newspapers come in various sizes, broadsheet being the
  largest, and tabloid being smaller, but what name is given to the smallest
  size? \textcolor{igsdCaseMuted}{[Reference: compact]}}\\
\multicolumn{2}{@{}p{\linewidth}@{}}{%
  \textbf{State:} The question distinguishes broadsheet from tabloid and asks
  for the smaller format \textcolor{igsdCaseMuted}{(format relation unresolved).}}\\
\midrule
\textbf{Student-sampled action} & \textbf{Hindsight-teacher proposal}\\
\textbf{Search:} \igsdStudent{Newspapers come in various sizes, broadsheet
being the largest, and tabloid being smaller, but what name is given to the
smallest size}
&
\textbf{Search:} \igsdTeacher{what is the name of the smallest size of
newspapers, where broadsheet is the largest and tabloid is smaller}\\
\textbf{Retrieved:} A tabloid is a newspaper with a compact page size smaller
than broadsheet. Some small-format papers refer to themselves as
\textbf{compact} newspapers \ldots{}
&
\textbf{Retrieved:} A tabloid is a newspaper with a compact page size smaller
than broadsheet \ldots{} The query remains anchored on the already-mentioned
\textbf{tabloid} format.\\
\textcolor{igsdCaseMuted}{\emph{Preserves the contrast among newspaper formats
and identifies the requested smaller format.}}
&
\textcolor{igsdCaseMuted}{\emph{Retrieves the tabloid page but treats the
question's tabloid reference as the answer.}}\\
\textbf{Answer:} \igsdCorrect{compact} --- correct
& \textbf{Answer:} \igsdIncorrect{tabloid} --- incorrect\\
\multicolumn{2}{@{}p{\linewidth}@{}}{%
  $\Delta\mathrm{IG}=\textcolor{igsdCaseBad}{-1.0606<0}$}\\
\bottomrule
\end{tabularx}
\endgroup
\end{table}

\end{document}